\documentclass[11pt,a4paper]{article}

\usepackage[utf8]{inputenc}
\usepackage[T1]{fontenc}
\usepackage{mathptmx}                   
\usepackage{helvet}                     
\usepackage{courier}
\usepackage{microtype}                  
\usepackage{textcomp}

\usepackage[margin=1in,headheight=22pt,includehead]{geometry}
\usepackage{setspace}
\usepackage{amsmath,amssymb,amsthm,amsfonts}
\usepackage{mathtools}
\usepackage{bm}

\usepackage{graphicx}
\usepackage{booktabs}
\usepackage{multirow}
\usepackage{makecell}
\usepackage{array}
\usepackage{float}
\usepackage{caption}

\usepackage[ruled,vlined,linesnumbered]{algorithm2e}
\SetKwInput{KwInput}{Inputs}
\SetKwInput{KwOutput}{Return}
\SetKw{KwTo}{to}
\SetAlCapSkip{0.6em}
\SetAlCapFnt{\normalfont\small\bfseries}
\SetAlCapNameFnt{\normalfont\small\itshape}
\SetAlgoCaptionSeparator{\,:\ }

\usepackage{xcolor}
\definecolor{arxivblue}{RGB}{26,54,104}        
\definecolor{arxivblue2}{RGB}{60,95,160}       
\definecolor{linkcol}{RGB}{35,80,150}          
\definecolor{abstractrule}{RGB}{180,190,210}   
\definecolor{rulecol}{RGB}{120,140,180}        

\usepackage[
    colorlinks=true,
    linkcolor=linkcol,
    citecolor=linkcol,
    urlcolor=linkcol,
    bookmarksnumbered=true
]{hyperref}

\usepackage{titlesec}
\titleformat{\section}
    {\normalfont\Large\bfseries\color{arxivblue}}
    {\thesection}{0.8em}{}
\titleformat{\subsection}
    {\normalfont\large\bfseries\color{arxivblue2}}
    {\thesubsection}{0.7em}{}
\titleformat{\subsubsection}
    {\normalfont\normalsize\bfseries}
    {\thesubsubsection}{0.6em}{}
\titlespacing*{\section}{0pt}{18pt}{8pt}
\titlespacing*{\subsection}{0pt}{14pt}{6pt}
\titlespacing*{\subsubsection}{0pt}{10pt}{4pt}

\usepackage{enumitem}
\setlist{itemsep=2pt,topsep=4pt,parsep=0pt}

\usepackage{fancyhdr}
\renewcommand{\headrule}{%
    \hbox to\headwidth{\color{rulecol}\leaders\hrule height 0.5pt\hfill}}

\fancypagestyle{plain}{%
    \fancyhf{}
    \fancyfoot[C]{\small\thepage}
    
    \renewcommand{\headrule}{}
}

\usepackage{titling}
\pretitle{\begin{center}\Large\bfseries\color{arxivblue}}
\posttitle{\par\vspace{6pt}\end{center}}
\preauthor{\begin{center}\normalsize}
\postauthor{\par\end{center}}
\predate{\begin{center}\footnotesize\itshape}
\postdate{\par\end{center}\vspace{0.3cm}}

\renewenvironment{abstract}
 {\vspace{2pt}
  \noindent\rule{\linewidth}{0.6pt}\vspace{-4pt}
  \par\noindent\textbf{\color{arxivblue}Abstract}\quad\ignorespaces}
 {\par\vspace{-2pt}\noindent\rule{\linewidth}{0.6pt}\par\vspace{10pt}}

\renewcommand{\arraystretch}{1.15}

\title{StyleVAR: Controllable Image Style Transfer via\\
       Visual Autoregressive Modeling\thanks{Independent work completed at Duke University.}}
\author{%
    Liqi Jing$^{1}$ \quad Dingming Zhang$^{1}$ \quad Peinian Li$^{1}$ \quad Lichen Zhu$^{1}$ \quad Yang Xu$^{2}$ \quad Hanyu Xing$^{3}$ \\[6pt]
    \normalsize $^{1}$Duke University \quad $^{2}$University of Southern California \quad $^{3}$Xidian University \\
    \normalsize \texttt{liqi.jing@duke.edu}
}
\date{April 21, 2026}

\begin{document}
\maketitle
\thispagestyle{plain}

\begingroup
\renewcommand{\arraystretch}{1.2}
\vspace{-1cm}
{\centering
\begin{tabular}{@{}r@{\ \ }l@{}}
\raisebox{-0.22\height}{\includegraphics[height=1.05em]{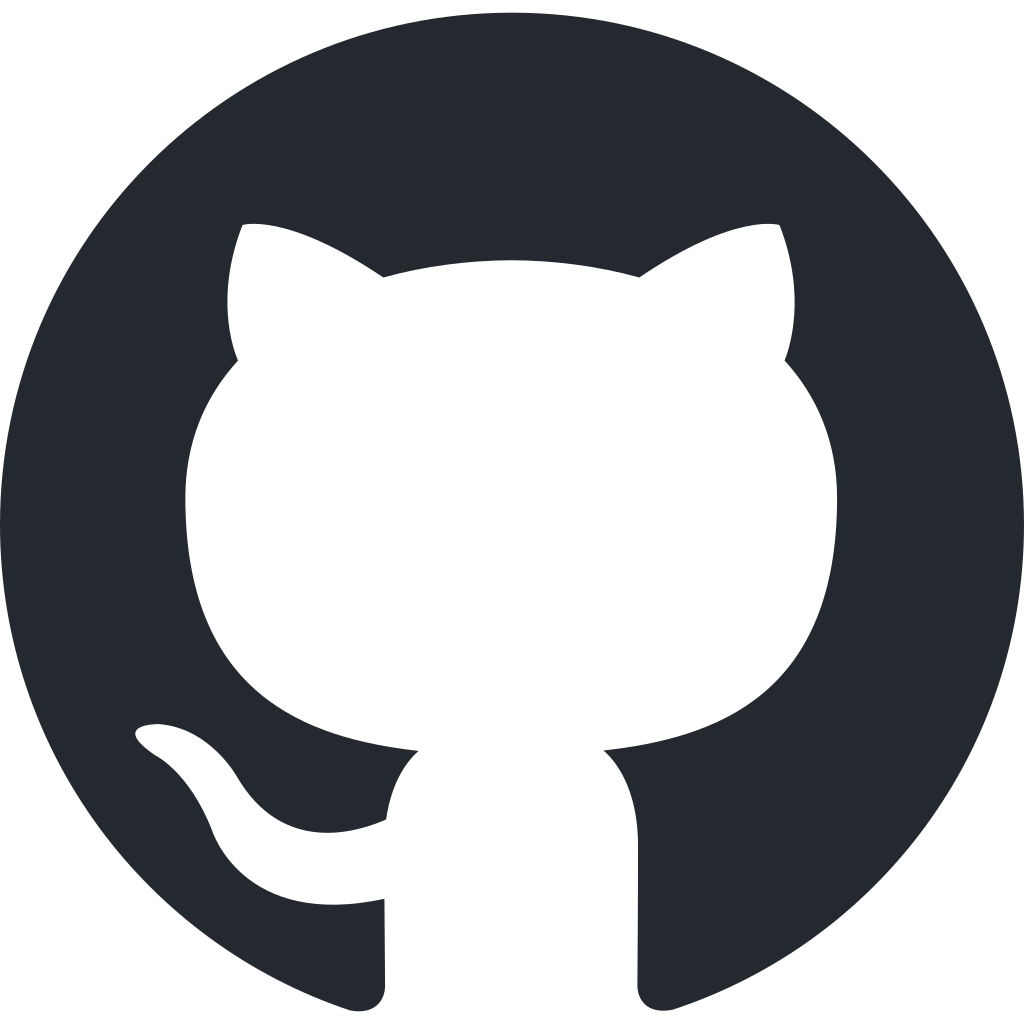}} \textbf{Code} &
  \href{https://github.com/Senfier-LiqiJing/StyleVAR}{\texttt{github.com/Senfier-LiqiJing/StyleVAR}} \\
\raisebox{-0.28\height}{\includegraphics[height=1.25em]{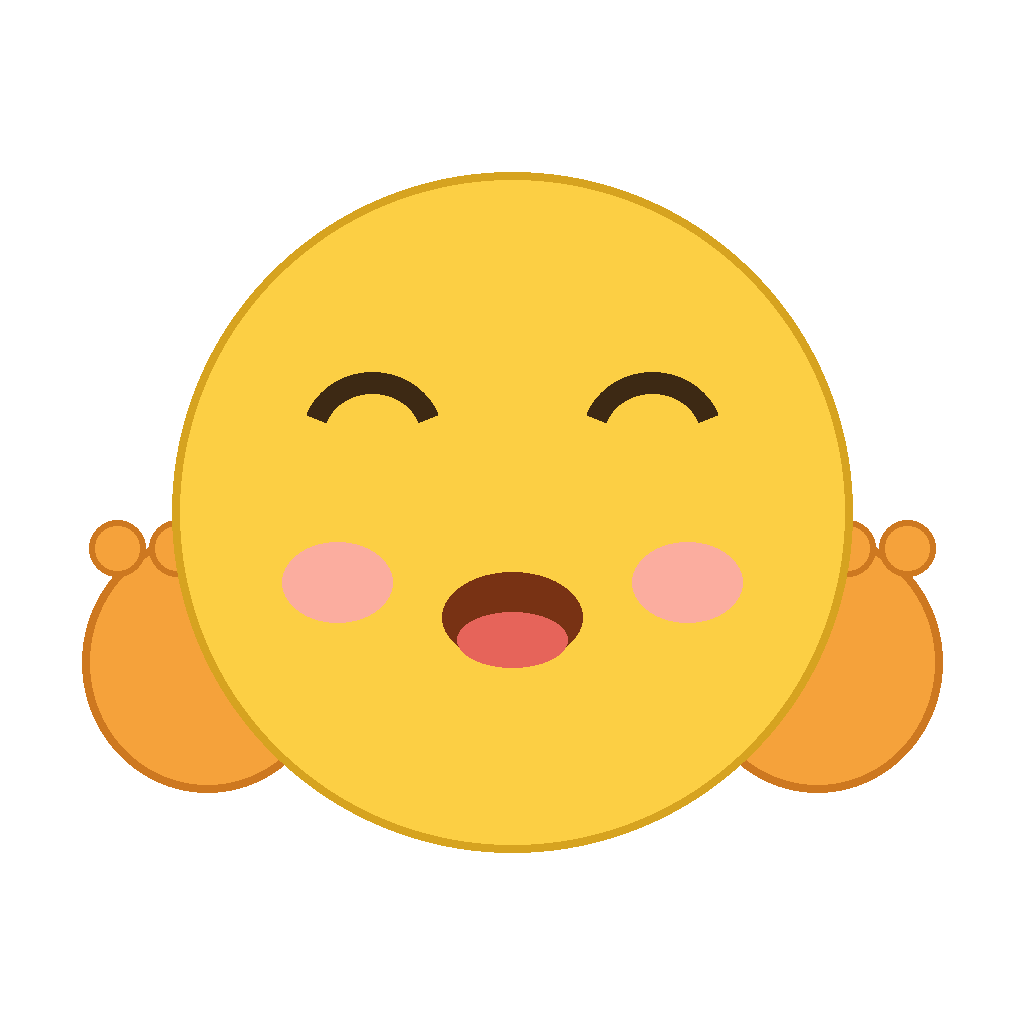}} \textbf{Weights} &
  \href{https://huggingface.co/Senfier-LiqiJing/StyleVAR}{\texttt{huggingface.co/Senfier-LiqiJing/StyleVAR}} \\
\end{tabular}\par}
\endgroup
\vspace{5pt}

\begin{abstract}
This project studies reference-based image style transfer: given a content image and a style image, the goal is to generate an output that preserves the structural semantics of the content while adopting the artistic texture of the style. We build on the Visual Autoregressive Modeling (VAR) framework and formulate style transfer as conditional discrete sequence modeling in a learned latent space. Images are decomposed into multi-scale representations and tokenized into discrete codes by a VQ-VAE; a transformer then autoregressively models the distribution of target tokens conditioned on style and content tokens. To inject style and content information, we introduce a blended cross-attention mechanism in which the evolving target representation attends to its own history, while style and content features act as queries that decide which aspects of this history to emphasize. A scale-dependent blending coefficient controls the relative influence of style and content at each stage, encouraging the synthesized representation to align with both the content structure and the style texture without breaking the autoregressive continuity of VAR. We train StyleVAR in two stages from a pretrained VAR checkpoint: supervised fine-tuning on a large triplet dataset of content--style--target images, followed by reinforcement fine-tuning with Group Relative Policy Optimization (GRPO) against a DreamSim-based perceptual reward, with per-action normalization weighting to rebalance credit across VAR's multi-scale hierarchy. Across three benchmarks spanning in-, near-, and out-of-distribution regimes, StyleVAR consistently outperforms an AdaIN baseline on Style Loss, Content Loss, LPIPS, SSIM, DreamSim, and CLIP similarity, and the GRPO stage yields further gains over the SFT checkpoint, most notably on the reward-aligned perceptual metrics. Qualitatively, the method transfers texture while maintaining semantic structure, especially for landscapes and architectural scenes, while a generalization gap on internet images and difficulty with human faces highlight the need for better content diversity and stronger structural priors.
\end{abstract}

\section{Introduction}
Reference-based image style transfer aims to generate an image that keeps what is in a content image while changing how it looks according to a style image. Concretely, the spatial layout and object semantics of the content image should be preserved, while colors, textures, and local patterns follow the chosen style. This setting is useful in artistic creation, visual prototyping, and controllable data augmentation, where users wish to restyle an existing scene without altering its high-level meaning. Achieving this objective requires a model that can respect content geometry and semantics while still expressing strong and diverse stylistic effects.

Balancing content preservation and style strength is challenging. If the model focuses too much on content, stylization becomes weak and the output resembles a slightly modified version of the original image. If the model overemphasizes style, it may distort object shapes or introduce artifacts that break semantic coherence. Styles also vary widely: some primarily change global tone, while others rely on fine-grained textures and patterns. A robust method must therefore provide a principled way to combine information from the content and style images, so that the resulting image is structurally faithful yet stylistically rich.

Earlier style transfer frameworks based on feed-forward CNNs and GANs showed that it is possible to learn powerful style priors, but they often come with practical limitations: models may need to be trained or adapted for specific styles, and training can be unstable or sensitive to dataset design. More recently, diffusion-based approaches have become the dominant backbone for high-quality image generation and have been adapted to style transfer as well. However, diffusion models typically require many iterative denoising steps, leading to slow sampling and high computational cost, and they often depend on additional guidance mechanisms or prompt-like controls that are not always ideal when we focus purely on image-to-image style transfer. These factors motivate exploring alternative formulations that retain strong generative capacity while offering better efficiency and more direct conditioning on content and style images.

In this project, we adopt the Visual Autoregressive Modeling (VAR) framework and cast style transfer as conditional discrete sequence modeling in a multi-scale latent space. Each image is decomposed into a hierarchy of feature maps and tokenized into discrete codes by a VQ-VAE encoder. The style image and content image are represented as sequences of tokens across scales, and the target image is generated scale by scale, with each set of target tokens conditioned on the history of previously generated tokens as well as the corresponding style and content tokens. This formulation explicitly encodes the intuition that the target should be consistent with its own past while being guided by both content structure and style appearance.

We train the resulting StyleVAR model in two stages. A supervised fine-tuning (SFT) stage optimizes a token-level cross-entropy loss on paired triplets and teaches the model to reproduce plausible stylizations, but leaves a gap between token-level supervision and the perceptual quality of the decoded image. We therefore add a second reinforcement fine-tuning stage based on Group Relative Policy Optimization (GRPO), using a DreamSim-based perceptual reward and per-action normalization weighting to counter the scale imbalance inherent to VAR's multi-scale hierarchy. Experiments across in-, near-, and out-of-distribution benchmarks show that StyleVAR substantially improves over an AdaIN baseline and that GRPO further sharpens the style/content trade-off learned during SFT.

\section{Related Works}

\subsection{Visual Autoregressive Modeling}

Autoregressive (AR) modeling has been the cornerstone of large language models, where text is generated sequentially, token by token, in raster-scan order. Extending this paradigm to images, however, is nontrivial: a single image contains tens of thousands of pixels or latent tokens, and the ordering used for text---left-to-right, top-to-bottom---does not respect the two-dimensional and multi-scale nature of visual content. Applying next-token AR directly to images therefore yields slow sampling, poor scaling behavior, and image quality that consistently trails behind diffusion models.

Visual Autoregressive Modeling (VAR)~\cite{var2024} reformulates image generation as \emph{next-scale prediction} rather than next-token prediction. Instead of flattening an image into a single long token sequence, VAR first encodes it with a multi-scale VQ-VAE into a hierarchy of token maps of increasing resolutions (e.g., $1{\times}1, 2{\times}2, \ldots, 16{\times}16$). A transformer then autoregressively predicts these token maps from coarse to fine, with all tokens within a given scale generated in parallel. This design aligns the generative order with the natural coarse-to-fine structure of visual perception: global layout is decided first at low resolutions, while local texture and detail are refined at higher resolutions.

The practical benefits of this formulation are substantial. On the ImageNet $256{\times}256$ benchmark, VAR improves Fr\'{e}chet Inception Distance (FID) from 18.65 to 1.73 over a conventional AR baseline, raises Inception Score from 80.4 to 350.2, and achieves roughly $20\times$ faster inference, surpassing diffusion transformers such as DiT in image quality, data efficiency, and scalability. VAR also exhibits scaling laws reminiscent of those observed in large language models, making it an attractive backbone for conditional visual generation. Our StyleVAR architecture is built directly on top of VAR, preserving its multi-scale token hierarchy and coarse-to-fine autoregressive process while augmenting the transformer with a blended cross-attention module that injects style and content conditions at each scale.

\subsection{Group Relative Policy Optimization}

Reinforcement learning from human or verifiable feedback has become a standard tool for aligning large generative models with downstream objectives that are difficult to express as differentiable losses. The dominant algorithm in this space has historically been Proximal Policy Optimization (PPO), which stabilizes on-policy updates through a clipped surrogate objective and relies on a learned value network (critic) to estimate advantages. In the context of large models, however, maintaining a separate critic of comparable capacity to the policy doubles the memory footprint and introduces additional training instability, since the critic itself must be kept calibrated throughout training.

Group Relative Policy Optimization (GRPO), proposed in DeepSeek-R1~\cite{deepseekr1}, removes the critic entirely. For each prompt, GRPO samples a group of $G$ candidate completions from the current policy, scores them with a reward function, and computes each trajectory's advantage by standardizing its reward against the mean and standard deviation of its group:
\begin{equation*}
    A^{(i)} = \frac{R^{(i)} - \mathrm{mean}_j R^{(j)}}{\mathrm{std}_j R^{(j)}}.
\end{equation*}
The group mean thus serves as a per-prompt baseline, eliminating the need for a learned value function while preserving variance reduction properties. The policy is then updated with a PPO-style clipped surrogate and a KL penalty toward a reference policy to prevent excessive drift. DeepSeek-R1 demonstrated that this simple formulation, combined with rule-based verifiable rewards, is sufficient to elicit sophisticated chain-of-thought reasoning in language models purely through RL, without supervised reasoning traces.

GRPO is especially attractive for visual generation, where candidate outputs can be scored by perceptual reward models and multiple stylizations of the same content--style pair are all plausibly valid. Group sampling naturally exploits this output diversity to construct informative advantages. We therefore adopt GRPO as the foundation of our reinforcement fine-tuning stage; its critic-free design is particularly beneficial given the already large memory footprint of VAR's multi-scale transformer.

\subsection{Credit Assignment for VAR}

Although GRPO transfers cleanly in spirit to visual autoregressive models, a direct application is suboptimal because VAR generation is structurally unlike language generation. In a language model, each step produces a single token over a fixed action space, so uniform credit assignment across the sequence is a reasonable default. In VAR, by contrast, each generation step produces a full token map whose size grows quadratically with the scale index: in the StyleVAR backbone, the scales range from a single token at $1{\times}1$ to 256 tokens at $16{\times}16$, yielding a $256\times$ imbalance in token count across steps. Sun et al.~\cite{varrl2026} characterize this as an \emph{asynchronous policy conflict}: different generation steps correspond to heterogeneous decision structures, and naively averaging the GRPO loss over all tokens lets the finest scales dominate the gradient even though the coarsest scales govern global layout and semantics.

To address this, \emph{VAR RL Done Right}~\cite{varrl2026} proposes three synergistic components on top of GRPO for VAR. First, a \emph{Value-as-Middle-Return} (VMR) intermediate reward decomposes the long-horizon RL problem at a middle timestep, providing denser and lower-variance feedback to early-stage generation. Second, \emph{Per-Action Normalization Weighting} (PANW) reweights the per-token loss by an inverse power of the scale's spatial resolution, normalizing the total gradient contribution of each scale and rebalancing credit across the multi-scale hierarchy; the authors find that a decay exponent of $\alpha \in [0.6, 0.8]$ is optimal. Third, a \emph{Mask Propagation} mechanism, inspired by Reward Feedback Learning, traces task-relevant spatial regions backward through the generation hierarchy and gates both intermediate rewards and gradients, focusing updates on causally relevant tokens.

Among these, PANW is directly applicable to any GRPO-style optimization of a multi-scale VAR model and is agnostic to the specific reward or task. We adopt PANW in our framework to restore meaningful gradient flow to the coarse scales that carry the structural decisions most critical for style transfer, while leaving scale-specific intermediate rewards and spatial masking to future work.

\section{StyleVAR: Architecture and Optimization}

\subsection{Blended Cross-Attention Autoregressive Modeling}

\paragraph{Formulation.} In the context of our style transfer task, the objective is to predict a target image that preserves the structural semantics of a content image $x_c$ while adopting the artistic texture of a style image $x_s$. Adopting the framework of Visual Autoregressive Modeling (VAR), we decompose images into multi-scale representations. Each scale's feature map is tokenized into discrete tokens. Formally, the style image tokens are denoted as $S = \{s_1, s_2, \dots, s_K\} = \mathcal{E}(x_s)$, and the content image tokens as $C = \{c_1, c_2, \dots, c_K\} = \mathcal{E}(x_c)$, where $K$ represents the total number of scales and $\mathcal{E}(\cdot)$ denotes the VQ-VAE tokenization process.

The generation of the target image, denoted as $R = \{r_1, r_2, \dots, r_K\}$, proceeds in a scale-wise autoregressive manner. Consequently, the autoregressive likelihood for StyleVAR is formulated as:
\begin{equation}
    \mathcal{P}(x\mid x_s,x_c) \;=\; \prod_{k=1}^K \mathcal{P}(r^k\mid r^{<k},s^k,c^k)
\end{equation}
where $r_k$ denotes the target features at the $k$-th scale, and $r_{<k} = r_{1:k-1}$ represents the history of generated target features prior to the $k$-th scale. Crucially, this formulation implies that the generation of the current scale is conditioned not only on the target's own history but also on the corresponding scale-specific features from the style and content conditions.

\begin{figure}[htbp]
    \centering
    \includegraphics[width=0.78\linewidth]{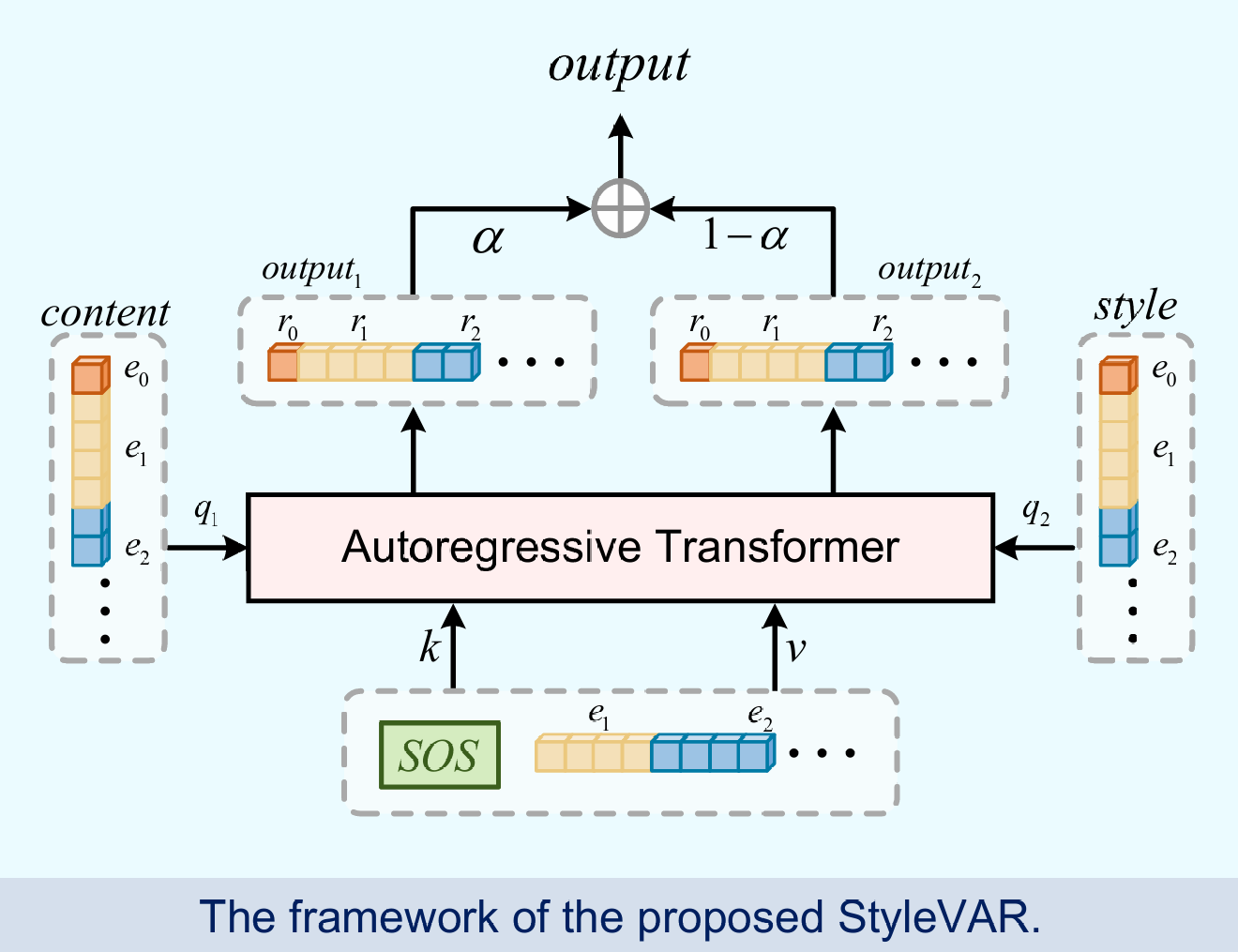}
    \caption{Structure of StyleVAR Transformers.}
    \label{fig:arch}
\end{figure}

\paragraph{Model Structure.} Figure~\ref{fig:arch} illustrates the architecture of StyleVAR. Building upon the VAR backbone, we introduce a Blended Cross-Attention mechanism to inject style and content information into the target image generation process. Within each transformer block, the feature update process is expressed as:
\begin{equation}
    h_{\text{new}} \;=\; h + \Big[\,\alpha \cdot \mathrm{Attn}(Q = s^k, K = h, V = h) \;+\; (1-\alpha) \cdot \mathrm{Attn}(Q = c^k, K=h, V=h)\,\Big]
\end{equation}
where $h$ represents the input target features at stage $k$ (or the output of the preceding transformer block). The term $\alpha_k$ is a heuristic hyperparameter governing the blending ratio between style and content information. Through this mechanism, $h$ passes through the transformer blocks and is iteratively updated via the injection of blended attention, ensuring the synthesized features align with both the content structure and style texture.

\subsection{Training and Inference}

\paragraph{Inference.} The inference process begins by initializing the start token at the first scale using content features extracted via a ResNet-18 backbone, which are projected to the embedding dimension via an MLP.

A critical component of the inference logic is the progressive accumulation of features. Unlike the generated target, the style and content features are fully observable; therefore, we pre-calculate their multi-scale ground truth tokens via VQ-VAE decomposition and accumulate them prior to inference. For the target image generation, we maintain a cumulative feature map, denoted as $\hat{f}$. At each step, the generated tokens $r_k$ are quantized and added to $\hat{f}$ in a residual manner. To serve as the input for the subsequent scale $k+1$, $\hat{f}$ is downsampled to the appropriate resolution. This downsampled map acts as the ``next-scale input,'' ensuring that the autoregressive generation maintains structural coherence across resolutions.

\paragraph{Training.} During the training phase, we employ a teacher-forcing strategy. The model concatenates the start token with the ground truth tokens of the target image across all other scales. Following the vanilla VAR paradigm, the model predicts the logits for stages $1$ to $K$ in parallel. We then calculate the Cross-Entropy loss between the predicted logits and the ground truth codebook indices. Note that, similar to inference, the input to the model at any stage $k$ is the accumulation of ground truth features from all preceding scales ($1$ to $k-1$), ensuring the model learns to refine coarse-grained features into fine-grained details.

\begin{algorithm}[!htbp]
\caption{StyleVAR Training}
\label{alg:training}
\KwInput{target image $x$, style image $x_s$, content image $x_c$;}
\KwInput{VQ-VAE Encoder $\mathcal{E}$, Codebook $Z$;}
\KwInput{Transformer $\mathcal{T}_\theta$ (parameters $\theta$);}
\tcp{1. Multi-scale Token Decomposition (Ground Truth)}
$R = \{r_1, \dots, r_K\} \leftarrow \text{VQVAE\_Encoding}(x)$\;
$S = \{s_1, \dots, s_K\} \leftarrow \text{VQVAE\_Encoding}(x_s)$\;
$C = \{c_1, \dots, c_K\} \leftarrow \text{VQVAE\_Encoding}(x_c)$\;
\tcp{2. Initialize Start Token}
$r_{start} \leftarrow \mathrm{MLP}(\mathrm{ResNet}(x_c))$\;
\tcp{3. Parallel Prediction (Teacher Forcing)}
$R_{input} \leftarrow [r_{start}, r_2, \dots, r_{K}]$ \tcp*{concatenate start token}
$L_{1:K} \leftarrow \mathcal{T}_\theta(R_{input}, S, C)$ \tcp*{Predict logits for all scales}
\tcp{4. Optimization}
$\mathcal{L} \leftarrow \mathrm{CrossEntropy}(L_{1:K}, R)$\;
$\theta \leftarrow \theta - \eta \nabla_\theta \mathcal{L}$\;
\KwOutput{Optimized parameters $\theta$;}
\end{algorithm}

\subsection{Discussion: Rationale for Attention Configuration}

\paragraph{Assignment of Query, Key, and Value.} A critical design decision in StyleVAR is the assignment of the target image features to the Key ($K$) and Value ($V$) roles, while assigning the style/content features to the Query ($Q$) role. This configuration diverges from standard cross-attention mechanisms where the target usually acts as the Query to retrieve information from the condition ($K, V$).

\paragraph{Preserving Autoregressive Continuity.} The efficacy of the vanilla VAR architecture hinges on its ``next-scale prediction'' paradigm. To generate the token map at stage $k$, the model requires a comprehensive aggregation of the entire history of generated tokens, $r_{<k} = r_{1:k-1}$, rather than solely relying on the immediate predecessor $r_{k-1}$. By designating the target feature history as $K$ and $V$, we ensure that the attention mechanism explicitly aggregates information from the target's own past. In this setup, the style and content features (as $Q$) act as a ``search query'', determining which parts of the target's history are most relevant to emphasize for the current generation step.

\begin{algorithm}[!htbp]
\caption{StyleVAR Inference (Autoregressive)}
\label{alg:inference}
\KwInput{style image $x_s$, content image $x_c$;}
\KwInput{Trained Transformer $\mathcal{T}_\theta$, VQ-VAE Decoder $\mathcal{D}$;}
\KwInput{Hyperparameters: steps $K$, resolutions $(h_k, w_k)_{k=1}^K$;}
\tcp{1. Prepare Conditions}
$S \leftarrow \text{VQVAE\_Encoding}(x_s)$\;
$C \leftarrow \text{VQVAE\_Encoding}(x_c)$\;
\tcp{2. Initialization}
$r_{input} \leftarrow \mathrm{MLP}(\mathrm{ResNet}(x_c))$\;
$\hat{f} \leftarrow 0$ \tcp*{Initialize cumulative feature map}
\tcp{3. Stage-wise Generation}
\For{$k = 1$ \KwTo $K$}{
    $Logits_k \leftarrow \mathcal{T}_\theta(r_{input}, s_k, c_k)$\;
    $r_k \sim \mathrm{Sample}(Logits_k)$ \tcp*{Top-k/Top-p sampling}
    $z_k \leftarrow \mathrm{lookup}(Z, r_k)$\;
    \tcp{Accumulate residual to max resolution}
    $\hat{f} \leftarrow \hat{f} + \mathrm{interpolate}(z_k, h_K, w_K)$\;
    \tcp{Prepare input for next scale}
    \If{$k < K$}{
        $r_{input} \leftarrow \mathrm{interpolate}(\hat{f}, h_{k+1}, w_{k+1})$\;
    }
}
\tcp{4. Image Reconstruction}
$\hat{x} \leftarrow \mathcal{D}(\hat{f})$\;
\KwOutput{Generated image $\hat{x}$;}
\end{algorithm}

\paragraph{Contrast with Alternative Configurations.} Conversely, if we were to assign the target features to $Q$ and the style/content to $K$ and $V$, the model would primarily attend to the external conditions ($s$ and $c$) to construct the next scale. While this maximizes information injection, it risks disrupting the autoregressive consistency fundamental to VAR. The model might rely too heavily on the static conditions and neglect the structural continuity required to evolve $r_{k-1}$ into $r_k$.

\paragraph{Theoretical Viability.} It is worth noting that by setting $V$ as the target features, the output of the attention block becomes a linear combination of the target's own history. While this does not directly ``copy'' pixels from the style image, the style-guided re-weighting (via the $Q \times K^T$ score) is theoretically sufficient to modulate the generative trajectory, effectively steering the autoregressive process to adopt the desired stylistic characteristics while maintaining the structural integrity of the target.

\subsection{GRPO-based Reinforcement Fine-tuning}

\paragraph{Motivation.} The supervised training described in Section~\ref{alg:training} optimizes a token-level cross-entropy loss against ground-truth codebook indices. While effective at teaching the model to reproduce training triplets, this objective leaves a gap between token-level supervision and the perceptual quality of the decoded image: predicting a slightly different codebook index at a fine scale may yield a nearly identical stylization, and conversely, being ``correct'' in token space does not guarantee perceptual fidelity to the target. Moreover, natural quality metrics for style transfer are applied to the decoded image and must pass through the non-differentiable VQ-VAE sampling step, making them unavailable as direct losses during SFT.

To bridge this gap, we introduce a second-stage reinforcement fine-tuning phase based on Group Relative Policy Optimization (GRPO). Reinforcement learning allows direct optimization of a perceptual reward computed on the decoded image, without requiring differentiability through the sampling trajectory. We choose GRPO over standard actor--critic formulations because it replaces the learned value function with a group-relative baseline: for each (content, style) condition, a group of trajectories is sampled, and each trajectory's advantage is computed relative to the group mean. This is particularly well suited to our setting --- style transfer admits many plausible stylizations for a given pair, and group sampling naturally exploits this diversity while avoiding the memory overhead of a separate critic network.

\paragraph{Reward Design.} We use DreamSim~\cite{dreamsim2023} between the generated image and the paired target image as the reward signal:
\begin{equation}
    R(\hat{x}, x) \;=\; -\lambda \cdot \mathrm{DreamSim}(\hat{x}, x),
\end{equation}
where $\hat{x}$ is the decoded generation, $x$ is the ground-truth target, and $\lambda$ is a scaling constant. DreamSim is a perceptual similarity metric calibrated against human judgments of mid-level image similarity. Compared to pixel-wise metrics or LPIPS --- which are dominated by low-level feature statistics --- and Gram-based style losses --- which capture only texture statistics and disregard semantic content --- DreamSim characterizes non-pixel-level image similarity, reflecting how close two images appear to a human observer in overall layout, content, and appearance. This makes it a more holistic training signal for a task in which both content preservation and stylistic alignment matter.

\paragraph{GRPO Formulation.} We cast the multi-scale autoregressive generation as a sequential decision process. At the $k$-th scale, the state $\mathfrak{s}_k$ consists of the accumulated target history $r_{<k}$ together with the scale-specific condition features $s^k$ and $c^k$, and the action $a_k$ is the set of discrete tokens produced at that scale. A trajectory $\mathbf{a} = \{a_1, \dots, a_K\}$ therefore spans all $K$ scales and is sampled autoregressively from the policy $\pi_\theta$.

For each training triplet $(x_c, x_s, x)$, we draw $G$ trajectories $\{\mathbf{a}^{(i)}\}_{i=1}^{G}$ from the rollout policy $\pi_{\theta_{\mathrm{old}}}$ and decode each one to an image $\hat{x}^{(i)}$. The group-relative advantage for trajectory $i$ is
\begin{equation}
    A^{(i)} \;=\; \frac{R^{(i)} - \mathrm{mean}_{j}\!\big(R^{(j)}\big)}{\mathrm{std}_{j}\!\big(R^{(j)}\big) + \varepsilon_{\mathrm{std}}},
\end{equation}
which normalizes each trajectory's reward against its peers and removes the need for a learned value baseline. Following PPO-style clipped optimization, we define the per-token importance ratio
\begin{equation}
    \rho_t^{(i)} \;=\; \frac{\pi_\theta\!\big(a_t^{(i)} \mid \mathfrak{s}_t^{(i)}\big)}{\pi_{\theta_{\mathrm{old}}}\!\big(a_t^{(i)} \mid \mathfrak{s}_t^{(i)}\big)},
\end{equation}
and the clipped surrogate
\begin{equation}
    \mathcal{L}^{\mathrm{PG}}_{t,i} \;=\; -\min\!\Big(\rho_t^{(i)} A^{(i)},\; \mathrm{clip}(\rho_t^{(i)}, 1-\epsilon, 1+\epsilon)\, A^{(i)}\Big).
\end{equation}
To prevent the policy from drifting too far from a reference policy $\pi_{\mathrm{ref}}$, we add a per-token KL penalty estimated via Schulman's unbiased, non-negative k3 estimator:
\begin{equation}
    \mathcal{L}^{\mathrm{KL}}_{t,i} \;=\; \exp\!\big(\log \pi_{\mathrm{ref}} - \log \pi_\theta\big) \;-\; \big(\log \pi_{\mathrm{ref}} - \log \pi_\theta\big) \;-\; 1.
\end{equation}

\paragraph{Scale-Aware Credit Assignment.} A naive GRPO implementation averages the per-token losses uniformly over the trajectory. This is problematic for VAR: the sequence length is dominated by the finest scales, with our 10 scales ranging from $1{\times}1$ (1 token) to $16{\times}16$ (256 tokens), a 256-fold imbalance. Under uniform averaging, the $16{\times}16$ scale alone would contribute 256 times more gradient than the $1{\times}1$ scale, yet coarse scales carry the global layout and semantic decisions that matter most for content preservation in style transfer. Gradient mass is therefore systematically misallocated toward high-frequency details.

To rebalance credit across scales, we adopt Per-Action Normalization Weighting (PANW)~\cite{varrl2026}. Each token at a scale of spatial resolution $h_k \times w_k$ is weighted by
\begin{equation}
    w_t \;=\; \frac{1}{Z}\,(h_k \cdot w_k)^{-\alpha},
\end{equation}
where $\alpha \in [0.6, 0.8]$ controls the strength of the rebalancing and $Z$ normalizes the weights to sum to one across the trajectory. We use $\alpha = 0.7$. Table~\ref{tab:panw} lists the resulting per-token weights; coarse scales receive roughly $60\times$ the per-token weight of the finest scale, which compensates for length imbalance and restores meaningful gradient at the scales that govern structure and layout.

\begin{table}[htbp]
    \centering
    \caption{Per-token PANW weights across the 10 VAR scales ($\alpha=0.7$). Scale denotes the spatial side length.}
    \label{tab:panw}
    \small
    \begin{tabular}{lcccccccccc}
        \toprule
        Scale ($h_k{=}w_k$) & 1 & 2 & 3 & 4 & 5 & 6 & 8 & 10 & 13 & 16 \\
        \# Tokens & 1 & 4 & 9 & 16 & 25 & 36 & 64 & 100 & 169 & 256 \\
        $w_t\,(\times 10^{-2})$ & 3.37 & 1.28 & 0.72 & 0.48 & 0.35 & 0.27 & 0.18 & 0.13 & 0.09 & 0.07 \\
        \bottomrule
    \end{tabular}
\end{table}

The final GRPO objective is a PANW-weighted sum of the policy and KL terms over all tokens of all trajectories in the group:
\begin{equation}
    \mathcal{L}_{\mathrm{GRPO}} \;=\; \frac{1}{G}\sum_{i=1}^{G}\sum_{t=1}^{L} w_t \,\Big(\mathcal{L}^{\mathrm{PG}}_{t,i} \;+\; \beta\, \mathcal{L}^{\mathrm{KL}}_{t,i}\Big),
\end{equation}
where $L$ is the total number of tokens in a trajectory (here $L=680$ across the 10 scales) and $\beta$ is the KL coefficient.

\paragraph{Iterative Reference Update.} We initialize the reference policy $\pi_{\mathrm{ref}}$ with the SFT model. In a standard reinforcement fine-tuning (RFT) pipeline, this reference is held fixed throughout RL training, so the KL term anchors the policy to the SFT distribution. As the policy improves, however, a static reference increasingly penalizes precisely the improvements we wish to keep. We therefore adopt an iterative variant: whenever the running average reward surpasses the reference's baseline by a meaningful margin, we replace the reference with the current policy, $\pi_{\mathrm{ref}} \leftarrow \pi_\theta$, and continue optimization against this improved anchor. This promotes our procedure from a single-shot RFT to iterative GRPO, enabling the policy to make sustained progress across many updates without being continually pulled back toward the original SFT distribution.

\begin{algorithm}[!htbp]
\caption{StyleVAR GRPO Fine-tuning}
\label{alg:grpo}
\KwInput{SFT-trained policy $\pi_\theta$; triplet dataset $\mathcal{D}$;}
\KwInput{group size $G$; clip ratio $\epsilon$; KL coefficient $\beta$; PANW exponent $\alpha$;}
\KwInput{reward $R(\hat{x}, x) = -\lambda \cdot \mathrm{DreamSim}(\hat{x}, x)$; update margin $\delta$;}
\tcp{1. Initialize reference policy with SFT weights}
$\pi_{\mathrm{ref}} \leftarrow \pi_\theta$; \quad $\bar{R}_{\mathrm{ref}} \leftarrow -\infty$\;
\For{each training step}{
    Sample triplet $(x_c, x_s, x) \sim \mathcal{D}$\;
    $\pi_{\theta_{\mathrm{old}}} \leftarrow \pi_\theta$ \tcp*{snapshot for on-policy ratio}
    \tcp{2. Group rollout}
    \For{$i = 1$ \KwTo $G$}{
        $\mathbf{a}^{(i)} \sim \pi_{\theta_{\mathrm{old}}}(\cdot \mid x_c, x_s)$\;
        $\hat{x}^{(i)} \leftarrow \mathrm{VAE\_Decode}(\mathbf{a}^{(i)})$\;
        $R^{(i)} \leftarrow R(\hat{x}^{(i)}, x)$\;
    }
    \tcp{3. Group-relative advantage}
    $A^{(i)} \leftarrow \big(R^{(i)} - \mathrm{mean}_j R^{(j)}\big) / \big(\mathrm{std}_j R^{(j)} + \varepsilon_{\mathrm{std}}\big)$\;
    \tcp{4. Teacher-forced log-probabilities on the sampled tokens}
    Compute $\log \pi_\theta(\mathbf{a}^{(i)})$, $\log \pi_{\theta_{\mathrm{old}}}(\mathbf{a}^{(i)})$, $\log \pi_{\mathrm{ref}}(\mathbf{a}^{(i)})$\;
    \tcp{5. PANW weights and GRPO loss}
    $w_t \leftarrow (h_k \cdot w_k)^{-\alpha}/Z$ for token $t$ at scale $k$\;
    Compute $\mathcal{L}_{\mathrm{GRPO}}$\;
    $\theta \leftarrow \theta - \eta\, \nabla_\theta \mathcal{L}_{\mathrm{GRPO}}$\;
    \tcp{6. Iterative reference update}
    \If{$\overline{R}_{\mathrm{recent}} > \bar{R}_{\mathrm{ref}} + \delta$}{
        $\pi_{\mathrm{ref}} \leftarrow \pi_\theta$; \quad $\bar{R}_{\mathrm{ref}} \leftarrow \overline{R}_{\mathrm{recent}}$\;
    }
}
\KwOutput{Fine-tuned policy $\pi_\theta$;}
\end{algorithm}

\section{Experiments}

\subsection{Implementation Details}

Our training proceeds in two stages: (1) supervised fine-tuning (SFT) of
StyleVAR on paired style-transfer data, and (2) reinforcement-learning
refinement via Group Relative Policy Optimization (GRPO) using a learned
perceptual reward.

\paragraph{Stage 1: Supervised Fine-Tuning.}
We initialized the weights of StyleVAR using the pre-trained vanilla VAR model.
To adapt the architecture for style transfer, we froze the VQ-VAE component
while fine-tuning the full 600M parameters of the transformer. Since StyleVAR
utilizes a dual-stream input (target features and content/style condition
features), the original projection layers of the vanilla VAR---responsible for
mapping image features to Query, Key, and Value (QKV)---were duplicated to
initialize the distinct projection layers for both the target and the
condition streams. The Feed-Forward Networks (FFN) were initialized with the
original VAR parameters. We fine-tuned the model for a total of 10 epochs. The
learning rate was scheduled at $5 \times 10^{-4}$ for the first 6 epochs and
decayed to $1 \times 10^{-4}$ for the final 4 epochs. Batchsize is set to be 128 with gradient accumulation. Training was conducted
on one NVIDIA 4090 (48GB) GPUs. 

\paragraph{Stage 2: GRPO Reinforcement Learning.}
After SFT, we further refine the policy on the same paired corpus by
optimizing the PANW-weighted GRPO objective of
Section~\ref{alg:grpo}. Because the transformer is large
($\sim$600M) and RL updates are noisy, we adopt LoRA
adapters (rank $r{=}256$, scaling $\alpha/r{=}2$) on every
attention ($W_Q^{\text{target}}, W_{QKV}^{\text{cond}}, W_{\text{proj}}$) and
FFN ($W_{\text{fc1}}, W_{\text{fc2}}$) linear layer, yielding 131M trainable
parameters (18.2\% of the backbone). The reference policy $\pi_{\mathrm{ref}}$
is realized \emph{in place} by disabling the LoRA path in a forward pass over
the same model, requiring no additional memory. For each content-style pair,
we sample $G{=}16$ autoregressive rollouts with top-$k{=}900$ and
top-$p{=}0.96$. The remaining GRPO hyperparameters are set to clip ratio
$\epsilon{=}0.2$, KL coefficient $\beta{=}0.1$, PANW exponent
$\gamma{=}0.7$, and reward scaling $\lambda{=}5.0$.

The iterative reference update of Section~3.4 is instantiated as a
\emph{peak-triggered iterative merge} on the LoRA adapter: when the reward
EMA exceeds the last-merge baseline by at least $\tau_{\text{gain}}{=}0.05$
and sustains this for $\tau_{\text{patience}}{=}50$ steps (after a minimum
cool-down of 300 steps), the LoRA delta is baked into the base weights and a
fresh zero-initialized adapter is attached, resetting the KL reference. An
emergency merge (shorter 50-step cool-down) is triggered whenever the raw KL
exceeds $2.0$ to prevent policy divergence.

We train with AdamW (lr $=1 \times 10^{-5}$, weight decay $=0.01$,
$(\beta_1,\beta_2){=}(0.9, 0.95)$), gradient clipping $1.0$, and full FP32
precision (mixed precision was found to cause $\log\pi_\theta$ drift between
rollout and update). GRPO training is carried out on a single NVIDIA 4090
(48GB) GPU with a physical batch size of 16 content-style pairs and
$G{=}16$ serial rollouts per pair.

\paragraph{Datasets.}
Both training stages use a concatenation of two paired style-transfer datasets:
\begin{itemize}
  \item \textbf{OmniStyle-150K}~\cite{omnistyle2025}: 143{,}992 (content, style,
  target) triplets covering a broad distribution of artistic styles over
  natural content.
  \item \textbf{ImagePulse}~\cite{ImagePulse-StyleTransfer2025}: 137{,}886 additional
  (content, style, target) triplets that enrich the stylization diversity
  and content domain coverage.
\end{itemize}
The two corpora are merged into a single 267{,}710-sample training set
(95\%/5\% train/val split). During SFT, we apply rotation and brightness
perturbations to content images and random cropping to style images, forcing
the network to learn fine-grained structural and textural details. GRPO
rollouts are performed without augmentation to keep the
content/style conditioning signal deterministic across the $G$ samples in a
group, which is necessary for the Z-score advantage to isolate policy
variance from input variance. For out-of-distribution evaluation we
additionally construct random (COCO, WikiArt) content-style pairs that break
the semantic correlation between the content and style images intentionally
present in the paired datasets.

\subsection{Results}

\paragraph{Quantitative Analysis.} We evaluate both StyleVAR checkpoints---the SFT model and the further GRPO-refined model---against the AdaIN baseline on three complementary benchmarks spanning in-distribution, near-distribution, and out-of-distribution regimes: (i) the OmniStyle-150K~\cite{omnistyle2025} held-out split (in-domain), (ii) the ImagePulse-StyleTransfer benchmark~\cite{ImagePulse-StyleTransfer2025} (near-domain), and (iii) a cross-dataset pairing of content images from MS-COCO~\cite{MS-COCO} with style images from WikiArt~\cite{WikiArt} (out-of-domain). For each benchmark we report six perceptual and structural metrics---Style Loss, Content Loss, LPIPS, SSIM, DreamSim~\cite{dreamsim2023}, and CLIP similarity---together with the per-sample inference cost.

Table~\ref{tab:main_results} summarizes the results. Both StyleVAR variants consistently outperform AdaIN on every quality-oriented metric across all three datasets, with the largest gains on SSIM (up to $+0.26$ on OmniStyle) and LPIPS (up to $-0.28$ on OmniStyle), indicating that the autoregressive multi-scale formulation preserves content structure far more faithfully than channel-wise feature statistics matching. The GRPO stage further improves over the SFT checkpoint on the majority of metrics on every dataset---most notably DreamSim and CLIP similarity, the two signals aligned with the reinforcement reward---confirming that reward-guided fine-tuning sharpens the style/content trade-off learned during SFT without destabilizing the policy. As expected, AdaIN retains a roughly two-orders-of-magnitude advantage in inference cost, reflecting the inherent gap between a single feed-forward normalization pass and a 10-scale autoregressive sampling procedure; closing this gap through distillation or parallel decoding is left to future work.

\begin{table}[htbp]
    \centering
    \caption{Cross-model and cross-dataset evaluation. We compare our StyleVAR under supervised fine-tuning (SFT) and GRPO reinforcement fine-tuning against the AdaIN baseline on the OmniStyle-150K, ImagePulse, and COCO$+$WikiArt benchmarks. Arrows indicate whether higher ($\uparrow$) or lower ($\downarrow$) is better. Best result within each dataset is in \textbf{bold}; second best is \underline{underlined}. Inference time is measured as seconds per generated sample on a single NVIDIA A100 (40GB).}
    \label{tab:main_results}
    \small
    \setlength{\tabcolsep}{4.5pt}
    \resizebox{\linewidth}{!}{%
    \begin{tabular}{l l ccccccc}
        \toprule
        \textbf{Dataset} & \textbf{Method}
            & \makecell{\textbf{Style Loss}\\$\downarrow$}
            & \makecell{\textbf{Content Loss}\\$\downarrow$}
            & \makecell{\textbf{LPIPS}\\$\downarrow$}
            & \makecell{\textbf{SSIM}\\$\uparrow$}
            & \makecell{\textbf{DreamSim}\\$\downarrow$}
            & \makecell{\textbf{CLIP Sim}\\$\uparrow$}
            & \makecell{\textbf{Infer (s)}\\$\downarrow$} \\
        \midrule
        \multirow{3}{*}{OmniStyle}
            & AdaIN (baseline)     & 0.0625           & 198.3449           & 0.7506           & 0.1421           & 0.6522           & 0.6555           & \textbf{0.0079} \\
            & StyleVAR (SFT)       & \underline{0.0468} & \underline{116.3569} & \underline{0.4743} & \underline{0.3975} & \underline{0.2276} & \underline{0.8704} & 0.4031 \\
            & StyleVAR (GRPO)      & \textbf{0.0466}  & \textbf{114.5686}  & \textbf{0.4656}  & \textbf{0.4024}  & \textbf{0.2164}  & \textbf{0.8740}  & 0.4031 \\
        \midrule
        \multirow{3}{*}{ImagePulse}
            & AdaIN (baseline)     & 0.0735           & 223.4699           & 0.7802           & 0.1574           & 0.6958           & 0.5651           & \textbf{0.0029} \\
            & StyleVAR (SFT)       & \underline{0.0452} & \textbf{180.7923}  & \underline{0.5618} & \underline{0.4282} & \underline{0.3168} & \underline{0.7903} & 0.4031 \\
            & StyleVAR (GRPO)      & \textbf{0.0387}  & \underline{182.0954} & \textbf{0.5572}  & \textbf{0.4320}  & \textbf{0.2979}  & \textbf{0.8000}  & 0.4031 \\
        \midrule
        \multirow{3}{*}{COCO$+$WikiArt}
            & AdaIN (baseline)     & 0.0282           & 171.0877           & 0.7688           & 0.1985           & 0.7536           & 0.5319           & \textbf{0.0027} \\
            & StyleVAR (SFT)       & \underline{0.0206} & \underline{160.1233} & \underline{0.7398} & \textbf{0.2713}  & \underline{0.6986} & \underline{0.5308} & 0.4031 \\
            & StyleVAR (GRPO)      & \textbf{0.0199}  & \textbf{157.5109}  & \textbf{0.7286}  & \underline{0.2677} & \textbf{0.6793}  & \textbf{0.5335}  & 0.4031 \\
        \bottomrule
    \end{tabular}%
    }
\end{table}

\begin{figure}[htbp]
    \centering
    \includegraphics[width=0.75\linewidth]{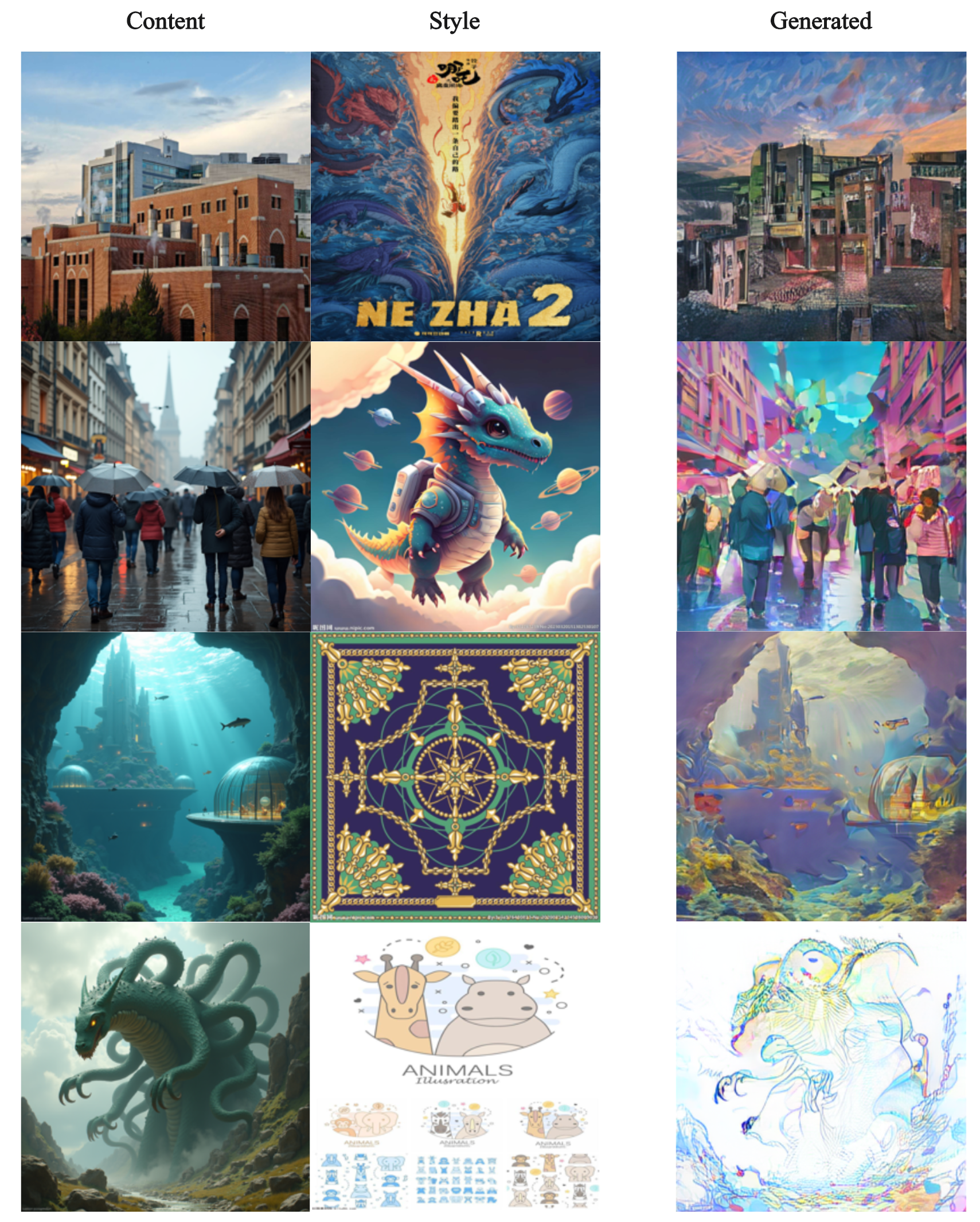}
    \caption{The generated images demonstrate that the model successfully transfers texture while maintaining the semantic structure of the content.}
    \label{fig:results}
\end{figure}

\paragraph{Qualitative Analysis.} To visually assess the model's performance, we generated samples using the validation set. Figure~\ref{fig:results} shows representative content--style--output triplets. The model successfully transfers color palettes and texture patterns while preserving the spatial layout and semantic content of the input scene.

\subsection{Limitations and Discussion}

Despite strong performance in the training and validation sets, our qualitative evaluation revealed a generalization gap when testing unseen images collected from the internet. This indicates a degree of overfitting to the training distribution. Upon further analysis of the OmniStyle-150K dataset, we identified a data imbalance: while the dataset contains about 150k triplets, these are generated from a limited pool of approximately 1{,}800 unique content images. Given StyleVAR's capacity (600M parameters), the model likely memorized the structural priors of this limited content set rather than learning a generalized representation of content structure.

\begin{figure}[htbp]
    \centering
    \includegraphics[width=0.82\linewidth]{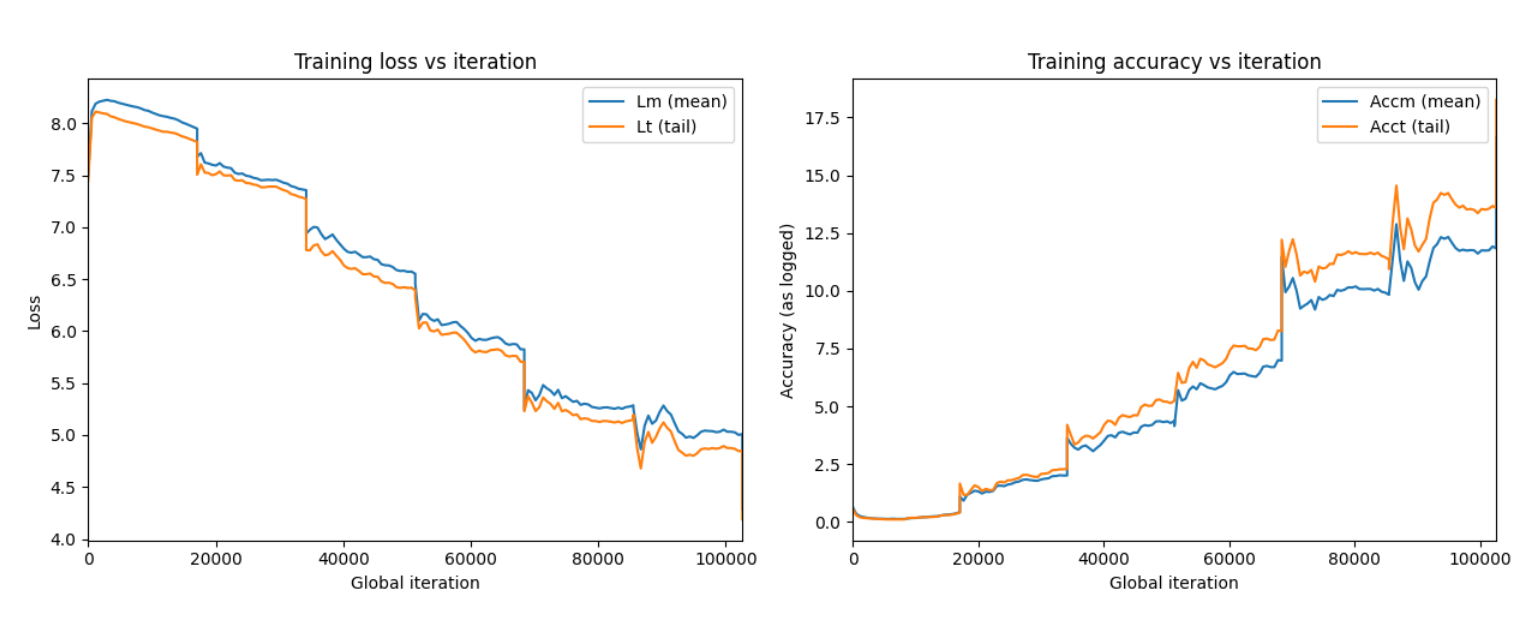}
    \caption{Loss and accuracy of training set across iterations.}
    \label{fig:loss}
\end{figure}
\begin{figure}[htbp]
    \centering
    \includegraphics[width=0.78\linewidth]{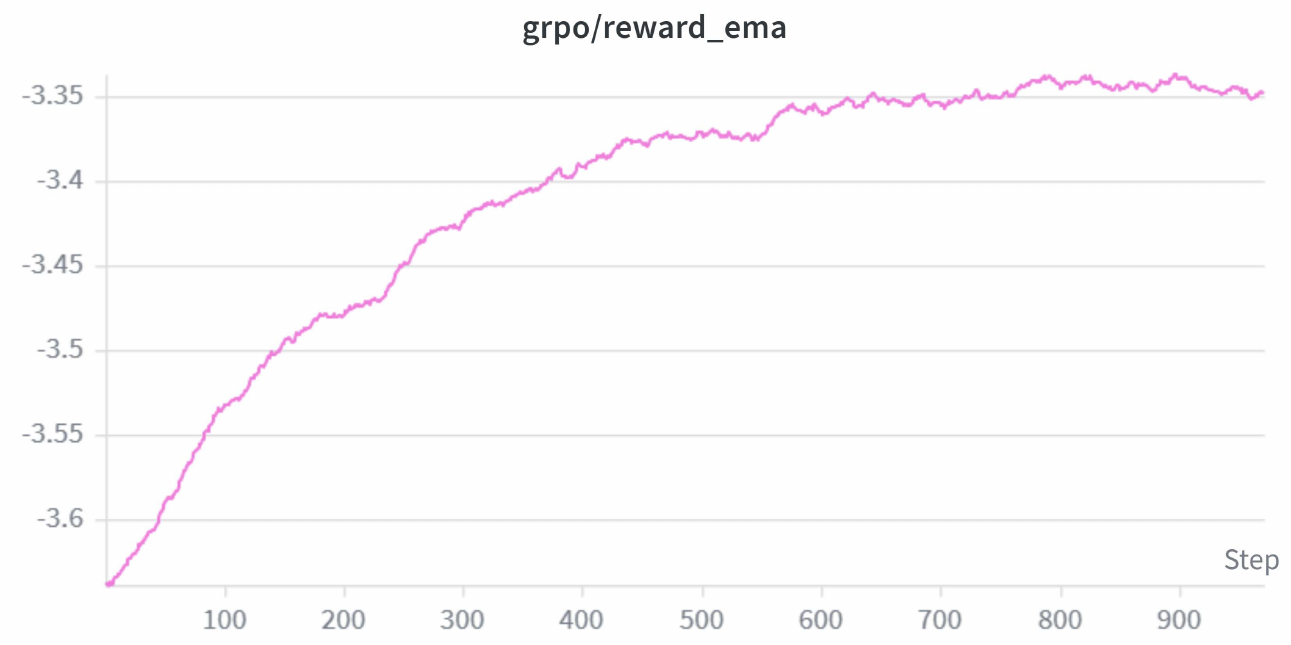}
    \caption{GRPO Reward-EMA across iterations.}
    \label{fig:grpo}
\end{figure}
Furthermore, we observed a performance disparity between different semantic domains. The model excels at stylizing landscapes and architectural scenes but struggles with human faces. This is likely due to two factors:
\begin{itemize}
    \item \textbf{Complexity.} Facial topology is significantly more complex and sensitive to structural deformation than natural scenes (e.g., mountains or buildings).
    \item \textbf{Perceptual Sensitivity.} Human visual perception is acutely sensitive to structural anomalies in facial features.
\end{itemize}

\section{Conclusion}
In this project, we presented StyleVAR, a framework that adapts Visual Autoregressive Modeling (VAR) to address the challenges of balancing content preservation and style intensity in reference-based image style transfer. By formulating the task as conditional discrete sequence modeling within a multi-scale latent space, we introduced a Blended Cross-Attention mechanism where style and content features function as Queries to selectively emphasize relevant aspects of the target's autoregressive history. On top of the supervised-fine-tuning stage, we further introduced a GRPO-based reinforcement fine-tuning stage that directly optimizes a DreamSim perceptual reward on the decoded image, with per-action normalization weighting to rebalance credit across VAR's scales. Experimental results demonstrate that StyleVAR generates stylistically rich and structurally coherent images, consistently outperforming the AdaIN baseline across all quality-oriented metrics on three in-, near-, and out-of-distribution benchmarks, and that the GRPO stage yields further gains over the SFT checkpoint on the majority of metrics---most notably the reward-aligned DreamSim and CLIP similarity. However, while the model excels in landscape and architectural scenes, qualitative analysis reveals a generalization gap on unseen internet images and difficulties with human faces, attributing these limitations to the lack of content diversity in the training dataset.

\end{document}